\documentclass[5p]{elsarticle}
\usepackage[T1]{fontenc}
\usepackage[utf8]{inputenc}

\usepackage{hyperref}

\usepackage{amsmath, amssymb}
\usepackage{xcolor}
\usepackage{graphicx,subcaption}
\usepackage{multirow}
\usepackage{enumitem}
\usepackage{booktabs}
\usepackage[ruled,vlined]{algorithm2e}

\usepackage{tikz}
\usetikzlibrary{shapes.geometric, arrows}
\usetikzlibrary{positioning, fit}

\tikzstyle{process} = [rectangle, rounded corners, minimum width=3cm, minimum height=1cm,text centered, draw=black, fill=gray!20]
\tikzstyle{data} = [rectangle, minimum width=3cm, minimum height=1cm, text centered, draw=black]

\tikzstyle{arrow} = [thick,->,>=stealth]

\begin{document}
\begin{frontmatter}

\title{A general technique for the estimation of farm animal body part weights from CT scans and its applications in a rabbit breeding program}
    
\author[medicopus,mate]{\'Adam Cs\'oka}
\ead{csoka.adam@sic.medicopus.hu}

\author[am]{György Kov\'acs}
\ead{gyuriofkovacs@gmail.com}

\author[mate]{Vir\'ag \'Acs}
\ead{acs.virag@uni-mate.hu}

\author[mate]{Zsolt Matics}
\ead{matics.Zsolt@uni-mate.hu}

\author[mate]{Zsolt Gerencs\'er}
\ead{gerencser.zsolt@uni-mate.hu}

\author[mate]{Zsolt Szendr\"o}
\ead{szendro.zsolt@uni-mate.hu}

\author[mate]{Istv\'an Nagy}
\ead{nagy.istvan.prof@uni-mate.hu}

\author[medicopus,mate]{\"Ors Petneh\'azy}
\ead{petnehazy.ors@sic.medicopus.hu}

\author[medicopus,kmok]{Imre Repa}
\ead{repa.imre@sic.medicopus.hu}

\author[kmok]{Mariann Moizs}
\ead{moizs.mariann@kmmk.hu}

\author[medicopus,mate]{Tam\'as Donk\'o\corref{cor0}}
\ead{donko.tamas@sic.medicopus.hu}

\address[medicopus]{Medicopus Nonprofit Ltd., Guba S. str. 40., Kaposv\'ar, 7400, Hungary}
\address[am]{Analytical Minds Ltd., \'Arp\'ad \'ut 5., Beregsur\'any, 4933, Hungary}
\address[mate]{Hungarian University of Agricultural and Life Sciences, Kaposv\'ar Campus, Guba S. str. 40., Kaposv\'ar, 7400, Hungary}
\address[kmok]{Moritz Kaposi General Hospital, Talli\'an Gy. str. 20-32., Kaposv\'ar, 7400, Hungary}

\cortext[cor0]{Corresponding author}

\begin{abstract}
Various applications of farm animal imaging are based on the estimation of weights of certain body parts and cuts from the CT images of animals. In many cases, the complexity of the problem is increased by the enormous variability of postures in CT images due to the scanning of non-sedated, living animals. In this paper, we propose a general and robust approach for the estimation of the weights of cuts and body parts from the CT images of (possibly) living animals. We adapt multi-atlas based segmentation driven by elastic registration and joint feature and model selection for the regression component to cape with the large number of features and low number of samples. The proposed technique is evaluated and illustrated through real applications in rabbit breeding programs, showing $r^2$ scores $12\%$ higher than previous techniques and methods that used to drive the selection so far. The proposed technique is easily adaptable to similar problems, consequently, it is shared in an open source software package for the benefit of the community.

\end{abstract}


\end{frontmatter}


\section{Introcution}
\label{intro}
 In the last decades, imaging techniques such as X-ray, ultrasound, magnetic resonance imaging, dual energy X-ray and computed tomography (CT) became gradually available for agricultural research and applications \citep{Scholz2015}. Due to the volumetric and geometrically accurate radiodensity images and cost effective operation, CT is a popular choice to estimate the body composition of living animals. The estimation of the weight or volume of tissues, organs and body parts enable meat quality predictions on different livestock animals, such as pig, beef and sheep \citep{Font-i-Furnols2013, Scholz2015}, and can drive breeding selection programs for various species, such as pig, rabbit and sheep \citep{Scholz2015, Gyovai2013, Matics2020}. 

To estimate the weights of various body parts from the CT images of living animals, the segmentation of the region is needed and usually the histogram of radiodensity values is regressed to reference weights to gain a predictive model that can be applied to the scans of non-dissected animals. We highlight that usually weights of the industrially valuable post-mortem cuts are of interest. These cuts are subject to the loss of liquids and truncation. Therefore, the reference weights used in regression are usually coming from dissection studies and the regression is expected to model all biases of the dissection process \cite{Ho2019, Gangsei2016}.

In the earliest attempts, the limited speed of CT imaging implied that the estimation of the amount of targeted muscles had to be carried out from a handful of cross-sectional images, acquired at predetermined anatomical locations. Usually these images were segmented manually and the sizes of the segmented regions were regressed to the weights measured during dissection \citep{Scholz2015}.
In the early 2000s, serial examinations became available, where sequences of images covering the whole body or a part of it could be made. Due to the large number of cross-sections, the manual processing became intractable.
The automated segmentation techniques proposed for various problems follow four main approaches: thresholding, statistical models, classical image processing pipelines and atlas techniques. Firstly, thresholding was proposed (\citep{Gyovai2013, Nagy2006, Vester-Christensen2009, Font-i-Furnols2009, Kasza2020}). The radiodensity values acquired by CT images are standardized to the Hounsfield-scale, and the thresholds are commonly accepted to distinguish fat, meat and bone voxels. Thresholding is applied primarily when the region of interest (ROI) can be distinguished by its radiodensity and no irrelevant tissues of similar densities are present. The voxel counts or histograms of the thresholded regions are regressed to the weights measured in dissection to reduce the systematic biases of imaging and dissection (\citep{Gyovai2013, Xiberta2017}). 
Some automated techniques are based on the extraction of spatial features and machine learning based classification as described in (\citep{Larsen2000, Pan2021}) and successfully applied in (\citep{Vester-Christensen2009}), however, these techniques can hardly tackle possible anatomical and geometrical constraints. Problem-specific segmentation techniques based on classical image processing operations were found to be overly specialized and hard to generalize to other problems \citep{Matics2020}. 
Single atlas driven techniques were proposed in (\citep{Ho2019, Gangsei2016}) and successfully applied for the segmentation of carcasses (\citep{Ngo2016, Gangsei2016a}). In \citep{Gangsei2016}, the atlas is derived as an average carcass, based on a dedicated algorithm for the segmentation of pig skeletons to drive the registration. In \citep{Ho2019}, the atlas represented as a set of parametric volumes is based on one animal, since averaging multiple volumes is claimed to smooth the boundaries of body parts overly.

Despite weight estimation from CT has applications targetting various body parts of numerous species, most of the techniques proposed so far are hard to adapt to a certain problem, due to species-specific steps, constraints or the lack of a sufficiently large number of annotated scans.

In this paper, we propose a multi-atlas based technique for the weight estimation problem in farm animal imaging. The main difference compared to similar approaches in medicine is that without creating a high quality segmentation, the statistical descriptors of the ensemble of segmentations are directly fed into a feature selection driven regression. By ignoring non-reliable features from the non-representative regions of various atlases, feature selection is expected to increase the accuracy of predictions. In two applications driving rabbit breeding programs, we demonstrate that the proposed technique leads to more accurate predictions than the use of one single atlas or an average atlas and also outperforms problem-specific segmentation techniques. The generality of the method refers to the lack of any species-specific step. The minimum requirements to adapt the method to a particular problem is one single CT scan with a manual annotation of the ROI, and a dissection study quantifying the weight of the ROI in some animals with (unlabeled) CT scans. The ability to incorporate multiple manual labelings makes the technique adaptable to available resources: more labelings can lead to better results. Due to the generality of the proposed method, it has been released as an open source package available in the GitHub repository \url{http://github.com/gykovacs/maweight}.
For complete reproducibility, the images and source code related to rabbit breeding are shared in the repository \url{http://github.com/gykovacs/rabbit_ct_weights}.

The rest of the paper is organized as follows. In Section \ref{prob}, an explicit formulation of the problem we address is provided. In Section \ref{propmeth}, the proposed technique is described in detail. Numerical experiments are discussed in Section \ref{rabbit}, finally, conclusions are drawn in Section \ref{conclusions}.

\section{Problem formulation}
\label{prob}

In this section, we formulate the problem we address and illustrate it through applications in breeding selection programs for rabbits.

\subsection{Weight estimation from farm animal CT images}

CT images are volumetric arrays of the reconstructed linear attenuation coefficiencts of voxels, representing how easily the voxel can be penetrated by the X-ray beam, and linearly scaled to air and water (under normal pressure and room temperature) having -1000 and 0 values, respectively. This scale is usually referred to as the Hounsfield-scale with Hounsfield-units (HU). The Hounsfield-scale characterizes the radiodensity of the material, which is not equivalent to the mass density. Since imaging artefacts (like partial volume effect - PVE) alter the HU values in a non-linear way (also depending on the composition of the subject), generic radiodensity-mass transfer functions are not supplied to CT instruments. Even if these functions were available, \emph{the target of weight estimation is usually the weight of the commercially valuable cuts which are subject to the loss of liquids (bleeding out) and truncation (due to the industrial processes extracting cuts)}. Therefore, the commonly accepted practice in the field is that the ROI is segmented and the histograms of HU values in these regions are regressed to the results of dissection studies measuring the weights of the targeted body parts. The regressor is expected to reconstruct the radiodensity-mass transfer function implicitly and also to account for the systematic biases of imaging and dissection. \emph{The most crucial step in this process is the segmentation of the ROI in the CT scans.} 

Naturally, there is a huge body of literature on the analysis of CT images in human medicine and industry (mainly related to quality assessment on conveyor belts). Although these well-established techniques provide great inspiration in farm animal imaging, the problems are slightly different and adaptation to the goals of farm animal imaging is needed. In contrast to applications in human medicine and manufacturing, the weight estimation problems of living animals have the following common characteristics and requirements:
\begin{itemize}[itemsep=0pt]
\item [R1:] Medicine being patient-oriented, human CT segmentation techniques are usually semi-automated \citep{Swierczynski2018, Lee2019}, controlled and possibly refined by medical experts. Contrarily, in farm animal imaging, a large number of scans need to be evaluated to draw conclusions about populations, thus, full-automation is preferred: \emph{segmentation needs to be done at scale}.
\item [R2:] Unlike in automated industrial applications of CT \cite{industry0, industry1} where regular objects are scanned without motion artifacts, in farm animal imaging, (possibly living) animals are the subjects of the studies: as sedation is prohibited by law prior to human consumption and tight immobilization risks injuries, the variability in postures is significantly higher than that in human medicine. Consequently, \emph{the segmentation method needs to be roboust to the variability of the living and usually non-sedated animals}.
\item [R3:] Farm animal research and breeding programs usually focus on various body parts, tissues and cuts of many spieces, thus, the development of dedicated segmentation techniques (like the enormous amount of research invested in the segmentation of the liver from human CT images \cite{liver}) is not feasible: \emph{the modeling needs to be done at scale, the segmentation technique should be easily adaptable to various problems in the field}.
\item [R4:] In many cases the ROI is not an entire organ or tissue with strong contours, rather, the ROI is subject to constraints determined by the subsequent steps of industrial processing. For example, in rabbit breeding programs, the muscle tissue on the ventral side of the backbone is irrelevant, as it cannot be extracted efficiently (see Figure \ref{atlas-manually-segmented-mld}). \emph{Consequently, the segmentation solution should be able to handle both anatomical and geometrical constraints}.
\end{itemize}

Our goal is to develop a generally applicable methodology for the estimation of the weights of various cuts in farm animal imaging, fulfilling requirements R1-R4 to facilitate automation and reduce costs by reducing the need for spieces-specific solutions.

\subsection{A real application: breeding selection for rabbits}
\label{real}

The problem is illustrated and the solution is evaluated through applications in rabbit breeding selection programs. The significance of these programs is illustrated by a brief overview of the ongoing work in the field. Rabbit meat production in the world has increased between 1998 - 2017 by nearly twice and now the European marketshare takes 20\% of that (\citep {Trocino2019}). 
Recently, the number of productive herds has decreased, but the size of herds and the role of large integrations are showing an increasing trend. It is important for market leaders to meet consumer needs as much as possible. Therefore the selection and research programs are targeting the genetic growth of the reproduction, carcass and meat quality which have great impact on the production level and the efficiency \citep{Nagy2006, Dalle_Szendro2011, Matics2020, Kasza2020}. The quantitative characteristics of live animals collected by CT image analysis are essential for a successful and effective selection program.

\begin{figure}[t]
    \centering
    \begin{subfigure}[t]{\linewidth}
        \centering
        \includegraphics[width=\linewidth]{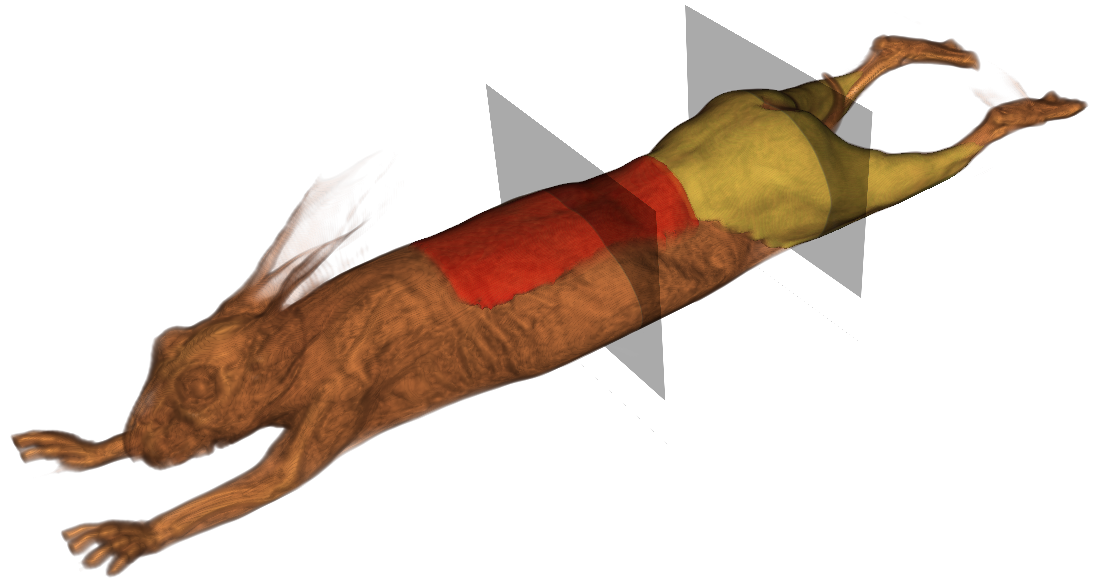}
        \subcaption{Atlas rabbit with manually segmented regions $($red: MLD; yellow: hinds$)$}
        \label{atlas-rabbit}
    \end{subfigure}
    \begin{subfigure}[t]{0.48\linewidth}
        \centering
        \includegraphics[width=\linewidth]{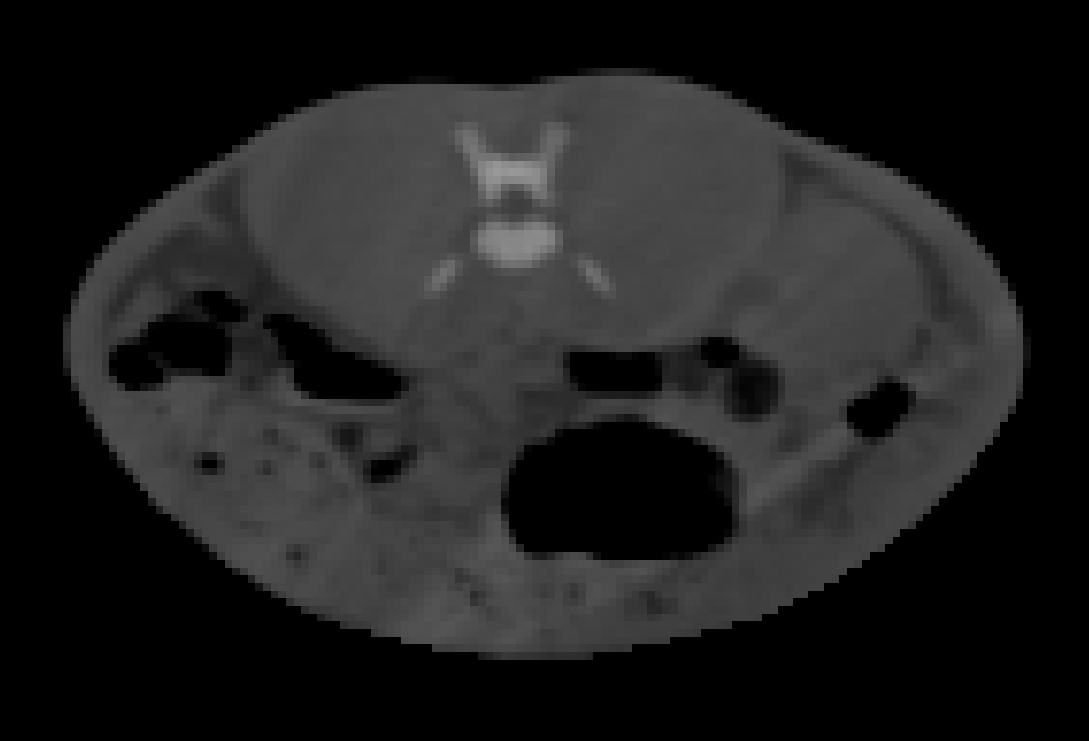}
        \caption{Cross-sectional slice at the MLD region}
        \label{atlas-mld}
    \end{subfigure}
    \begin{subfigure}[t]{0.48\linewidth}
        \centering
        \includegraphics[width=\linewidth]{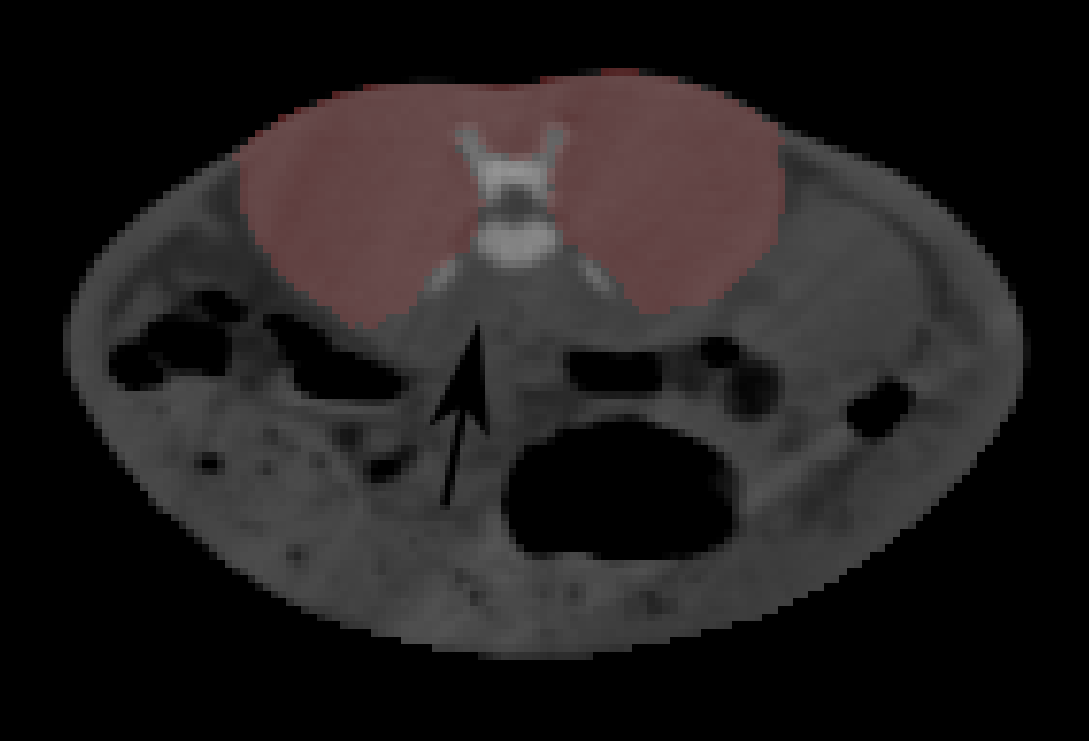}
        \caption{Manual segmentation in the MLD region}
        \label{atlas-manually-segmented-mld}
    \end{subfigure}
    \begin{subfigure}[t]{0.48\linewidth}
        \centering
        \includegraphics[width=\linewidth]{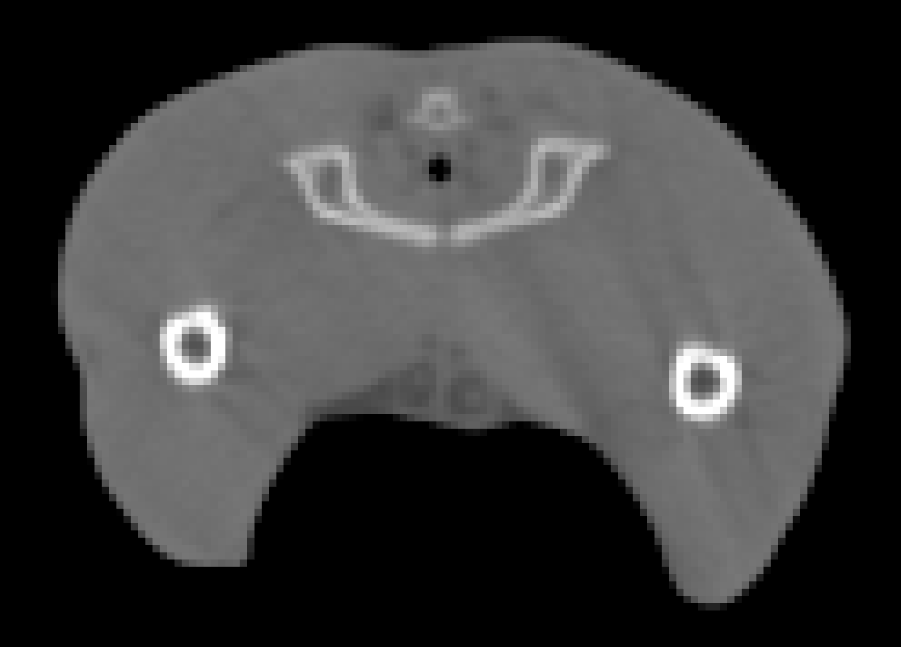}
        \caption{Cross-sectional slice at the hinds region}
        \label{atlas-hinds}
    \end{subfigure}    
    \begin{subfigure}[t]{0.48\linewidth}
        \centering
        \includegraphics[width=\linewidth]{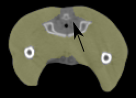}
        \caption{Manual segmentation in the hinds region}
        \label{atlas-manually-segmented-hinds}
    \end{subfigure}
    
    \caption{The illustration of the manual segmentations with cross-sectional slices. Geometric constraints are marked with arrows: these regions are connected to and have similar intensities as the manually annotated colored regions, but are not parts of the regions of interest.}
    \label{fig1}
\end{figure}

In the breeding selection program running at the Hungarian University of Agriculture and Life Sciences, Kaposvár Campus, Hungary, two regions of muscles are targeted by the program to be increased in weight:
the Musculus Longissimus Dorsi (MLD),
and the muscles in the hind region. For illustration, the manual labelings of these regions are denoted by red and yellow in Figure \ref{fig1}, respectively.
In Figures \ref{atlas-manually-segmented-mld} and \ref{atlas-manually-segmented-hinds} arrows indicate regions which are connected and homogenous to the ROI but are not targeted by the selection program. These examples illustrate the anatomical and geometrical constraints which are usually imposed in the class of problems we address. The estimation of weights of the two regions are treated as two separate problems as the industry processes these regions separately and they serve as separate inputs to the selection program.

\section{The proposed method}
\label{propmeth}

In this section, an overview of the proposed weight estimation technique is given and its steps are discussed in details. 
Although the images used to illustrate the steps of the proposed method are related to applications in rabbit breeding programs, we highlight that \emph{no steps are specific to this spieces}.

\subsection{Overview of the solution}
\label{propmeth-overview}

Instead of simply declaring the proposed method, first we discuss how the requirements phrased in Section \ref{prob} shaped the approach and the selection of techniques we adapt from the literature. 

As the target of the segmentation is not necessarily a homogenous and connected tissue, but can be subject to anatomical or geometrical constraints (R4), those techniques of human medicine which focus on the segmentation of homogenous regions cannot be exploited \citep{liver}. Probably the easiest way to represent these additional constraints without a codification overly specific to spieces (R3) (like constraints relative to certain characteristic points of the skeleton) is by examples, namely, manually labeling some scans. A sufficiently large number of manual annotations could enable the use of popular deep learning techniques \citep{Pan2021}. However, in practice, the manual labeling of CT images is an expensive and time consuming operation. It is unrealistic to expect more than a handful of labelings which is clearly insufficient for high capacity machine learning techniques like deep learning. Another way to exploit the knowledge encoded in a manual labeling is treating it as an atlas, registering the base image to new images and transform the manual labeling along to gain a segmentation. In this case, the registration technique needs to handle the enormous variability of the images (R2), hence, non-rigid, elastic registration techniques should be used. Given the well-established elastic registration techniques developed in decades of research \citep{elastix}, if the circumstances of the acquisition are sufficiently standardized - despite the loose immobilization of the subjects - one can expect the registration operating in a fully automated manner (R1). However, registration being a stochastic optimization, in certain cases it might end up in suboptimal solutions. To improve the robustness of the technique, multiple manual labelings (multiple atlases) can be exploited. Multi-atlas techniques \citep{survey} are well-established in human medicine, and commonly, the labelings derived from the individual atlases are unified into one single high quality segmentation for visual inspection by medical experts. For weight estimation, the unification of labelings is an informational bottleneck as the fine details of the individual segmentations are smoothed before regression. Feeding the statistical descriptors (like histograms) of the individual segmentations into regression lets the regression technique assign more weight to representative features and smaller weights to features describing regions of atlases unreliably. Finally, in order to let the regression deal with a relatively large number of features and a relatively low number of samples, strong regularization and preferrably feature selection is needed.

\begin{figure}[t]
\begin{footnotesize}
\begin{tikzpicture}[node distance=1.5cm, on grid]
\node (dissected-ct) [data, align=left] {CT scans of living \\ animals before \\ dissection};
\node (dissected-weights) [data, align=left, below of = dissected-ct] {Weights of regions\\ after dissection};
\node[rectangle, draw=black, dotted, fit=(dissected-ct) (dissected-weights), inner sep=2mm, label=above:Dissected set ($D$)] (all-dissected) {};

\node (manual-ct) [data, align=left, right =4.5cm of dissected-ct] {CT scans of living \\ animals};
\node (manual-weights) [data, align=left, below of = manual-ct] {Manual segmen-\\tation of regions};
\node[rectangle, draw=black, dotted, fit=(manual-ct) (manual-weights), inner sep=2mm, label=above:Manually segm. set ($M$)] (all-manual) {};

\node[rectangle, draw=black, dotted, fit=(all-dissected) (all-manual), inner xsep=3mm, inner ysep=5mm, label=above:Input] (all-input) {};

\node (regseg) [process, align=left, below = 3cm of all-input] {Segmentation of CT scans of dissected\\animals by registration};

\draw[arrow] (dissected-ct) -- ++(1.8,0) -| (regseg);
\draw[arrow] (manual-ct) -- ++(-1.8,0) -| (regseg);
\draw[arrow] (manual-weights) -- ++(-1.8,0) -| (regseg);

\node (segmentations) [data, align=left, below of=regseg] {Segmented masks};

\draw[arrow] (regseg) -- (segmentations);

\node (featext) [process, align=left, below of = segmentations] {Feature extraction};

\draw[arrow] (segmentations) -- (featext);

\node (training-set) [data, align=left, below of=featext] {Regression training set};

\draw[arrow] (featext) -- (training-set);
\draw[arrow] (dissected-weights) -- ++(-1.85, 0.0) -- ++(0.0, -6.5) |- (training-set);

\node (regr) [process, align=left, below of = training-set] {Regression model fitting with feature selection};

\draw[arrow] (training-set) -- (regr);

\node (regression) [data, align=left, below of = regr] {Prediction model};

\draw[arrow] (regr) -- (regression);

\end{tikzpicture}
\end{footnotesize}
\caption{Overview of the training phase of the proposed method, the steps with gray background are discussed in details in subsections \ref{segreg}, \ref{features}, \ref{modelselection}}
\label{model-fitting}
\end{figure}
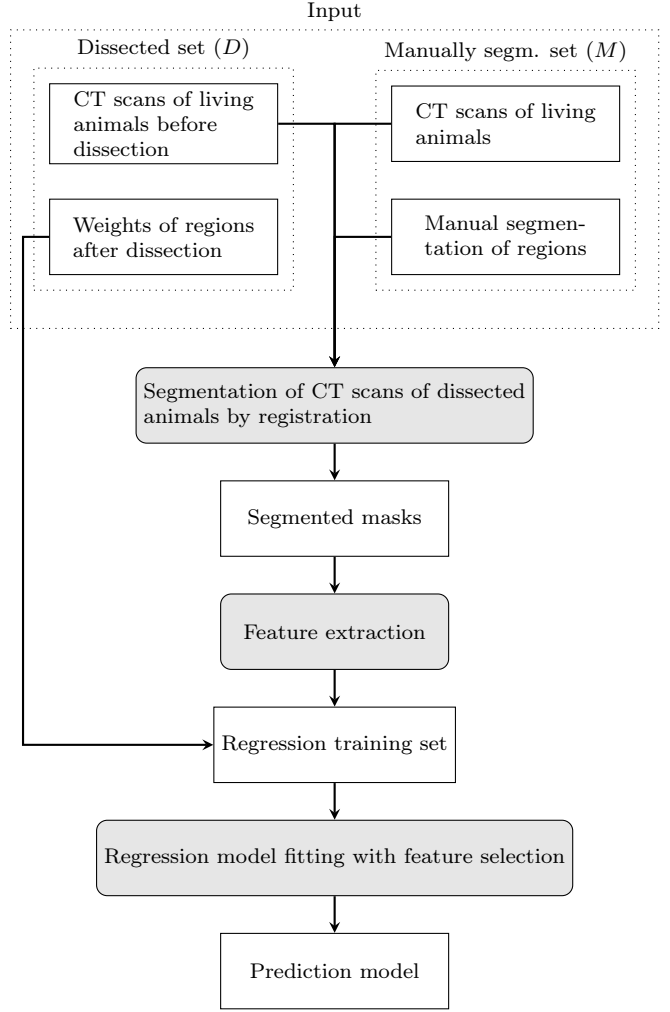

Based on these considerations, the flowchart of the proposed method is visualized in Figure \ref{model-fitting}. The input consists of two sets of CT images: set $M$ with manual segmentations, and set $D$ with corresponding weights of ROIs measured in  dissection studies ($D$). The first step is to segment the images of set $D$ by registration, using the manually segmented images from set $M$ as atlases. For each image in set $D$ this will result $M$ individual segmentation masks. Then, features describing the segmented regions are extracted, leading to a regression training set. The use of features from multiple, independent segmentation masks of an image is expected to increase the roboustness and descriptive power of the regressor. Finally, a predictive regression model is fitted with the dissection weights as target variables. 

The processing of unseen images follows the very same steps as the fitting of the prediction model: first, the manual segmentations of set $M$ are registered and transformed to the unseen image, then, statistical features are extracted from the ensemble of segmentations, and finally, the regression model is exploited to predict the weights based on the features. 

\subsection{Multi-atlas segmentation by elastic registration}
\label{segreg}

Atlas based segmentation techniques were developed gradually in the past 15 years, mainly for the segmentation of CT and MRI images of various human body parts and organs \cite{survey}. The main idea of atlas techniques is that given a high quality manual annotation (usually called the atlas) of some regions in CT image A, if one could find the transformation field transforming image A into image B, then the transformation field could be used to transform the manual annotation, too, leading to an annotation of image B. 
A detailed overview of related techniques can be found in the survey \cite{survey}. 

The transfomation field between two images is usually found by registration techniques. Due to the enormous variability of the CT images of (possibly) living animals, we exploit non-rigid (elastic) registration through the \emph{elastix} software \cite{elastix} providing a high level interface with roboust default parameter settings for the registration functionalities of the ITK package \cite{itk}, which is commonly treated as the \emph{de facto} implementation of registration techniques for human medicine. 

\begin{figure}[ht]
    \centering
    \begin{subfigure}[t]{0.48\linewidth}
        \centering
        \includegraphics[width=\linewidth]{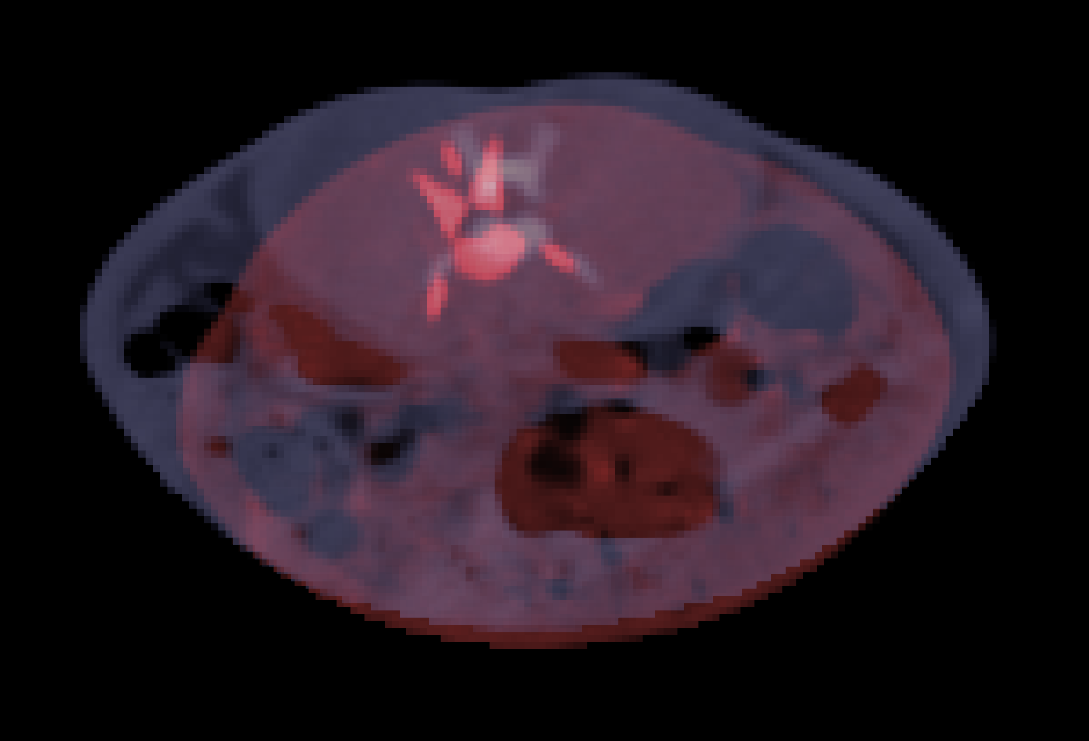}
        \subcaption{MLD overlay prior to registration}
        \label{mld-before-reg}
    \end{subfigure}
    \begin{subfigure}[t]{0.48\linewidth}
        \centering
        \includegraphics[width=\linewidth]{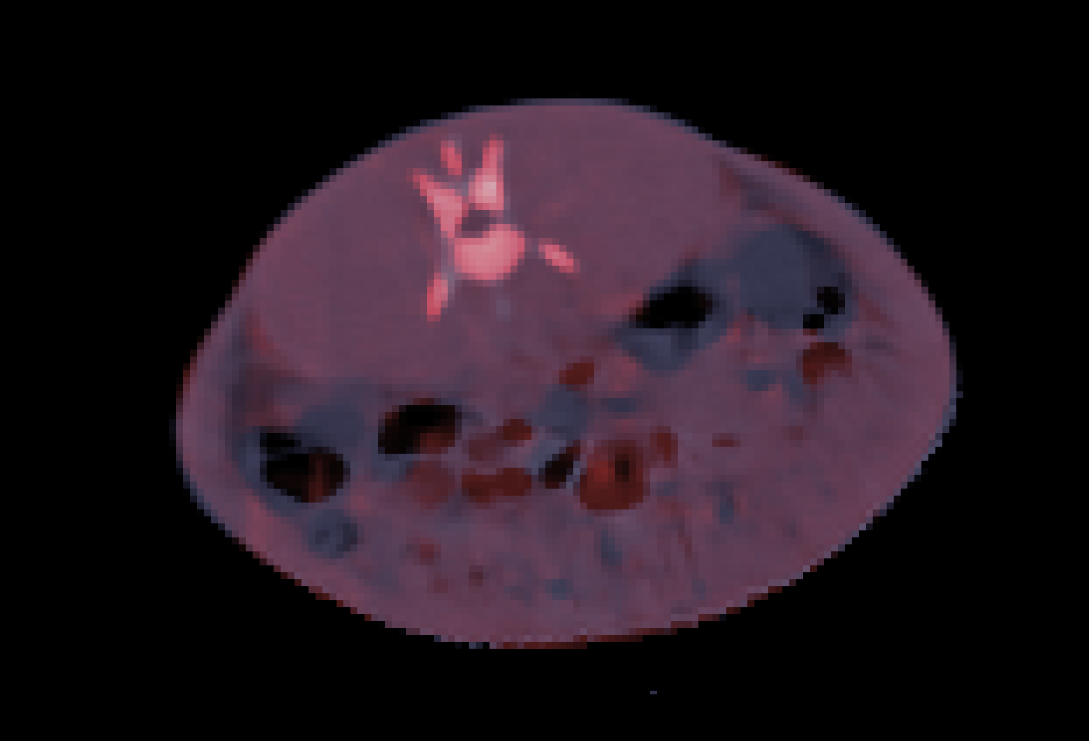}
        \caption{MLD overlay after registration}
        \label{mld-regged}
    \end{subfigure}
    
    \centering
    \begin{subfigure}[t]{0.48\linewidth}
        \centering
        \includegraphics[width=\linewidth]{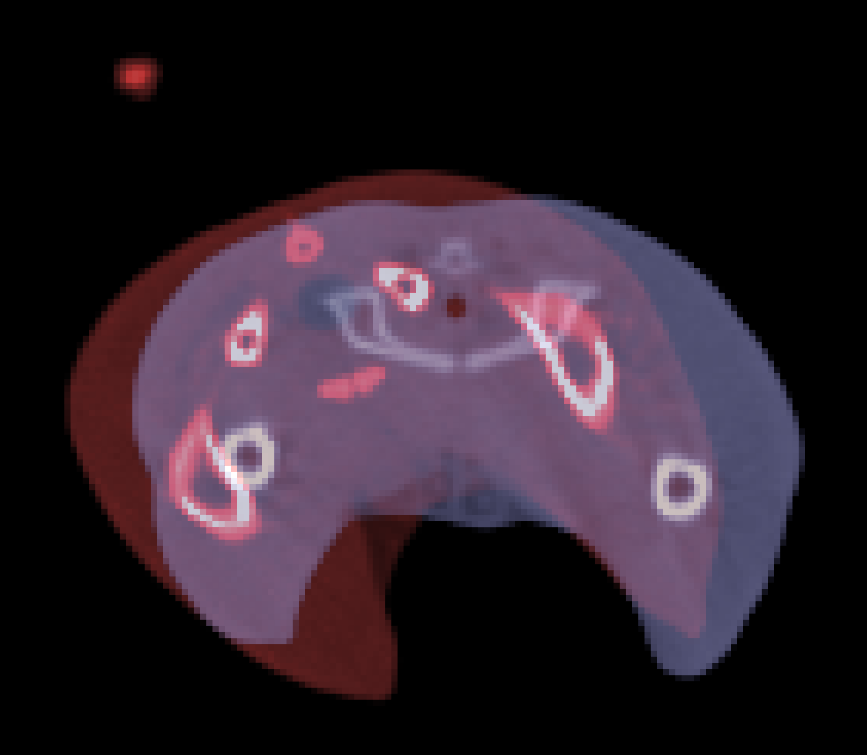}
        \subcaption{Hinds overlay prior to registration}
        \label{hinds-before-reg}
    \end{subfigure}
    \begin{subfigure}[t]{0.48\linewidth}
        \centering
        \includegraphics[width=\linewidth]{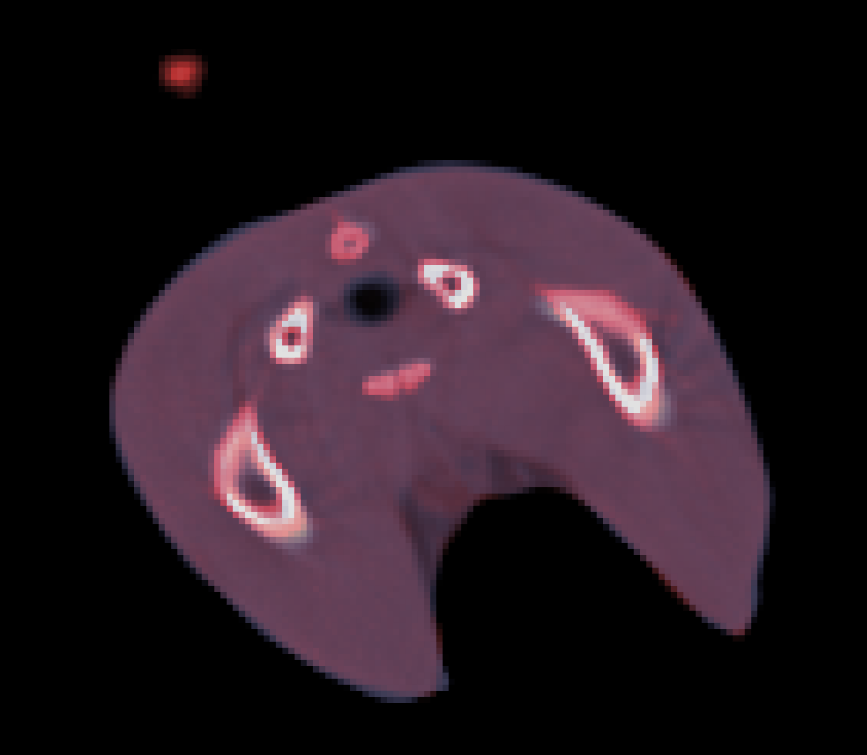}
        \caption{Hinds overlay after registration}
        \label{hinds-regged}
    \end{subfigure}
    \caption{MLD and hinds slices overlaid for a randomly selected image (red) and an atlas-image (blue), prior to and subsequent to the elastic registration.}
    \label{fig2}
\end{figure}
\begin{figure}[ht]
    \centering
    \begin{subfigure}[t]{\linewidth}
        \centering
        \includegraphics[width=\linewidth]{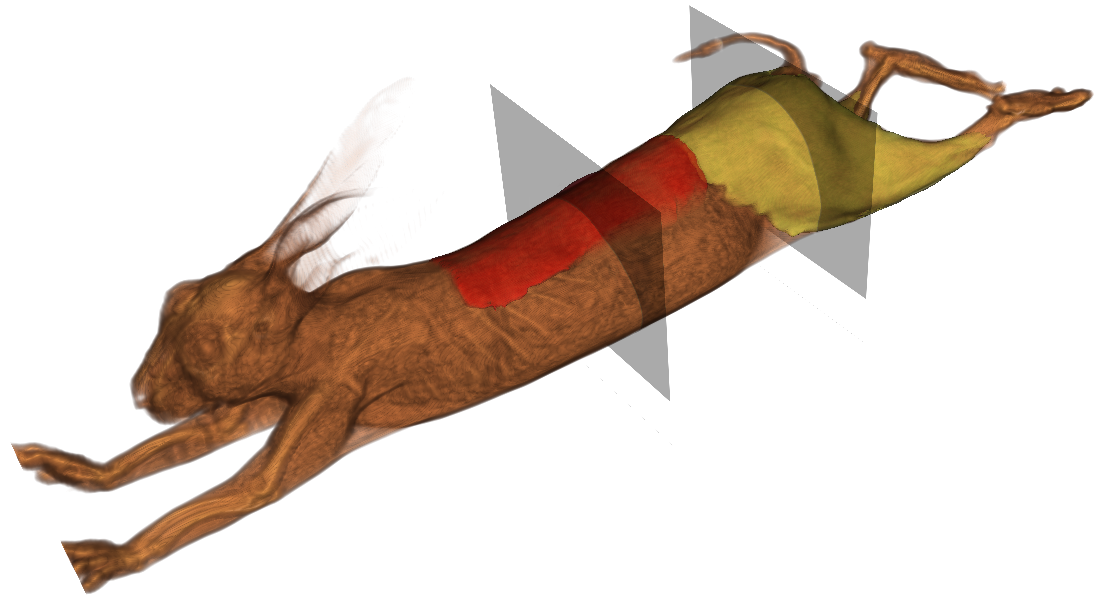}
        \subcaption{Unseen rabbit with registered regions $($red: MLD; yellow: hinds$)$}
        \label{unseenrabbit}
    \end{subfigure}
    \begin{subfigure}[t]{0.48\linewidth}
        \centering
        \includegraphics[width=\linewidth]{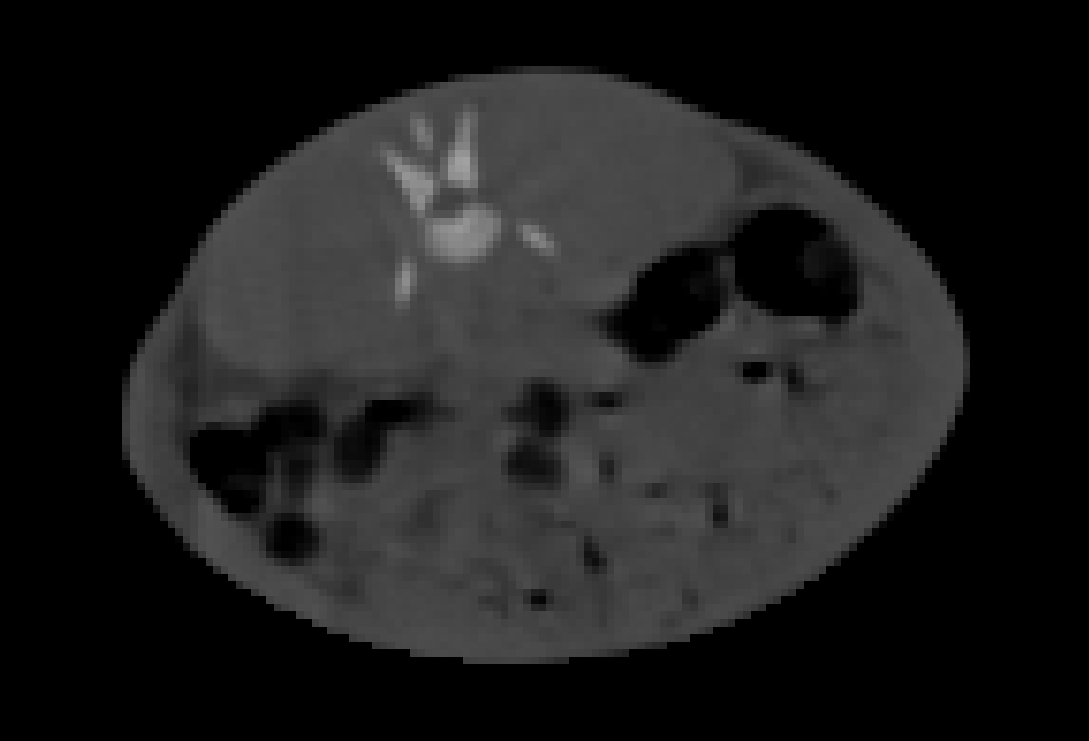}
        \caption{Cross-sectional slice at red region without segmentation}
        \label{unseen-mld}
    \end{subfigure}
    \begin{subfigure}[t]{0.48\linewidth}
        \centering
        \includegraphics[width=\linewidth]{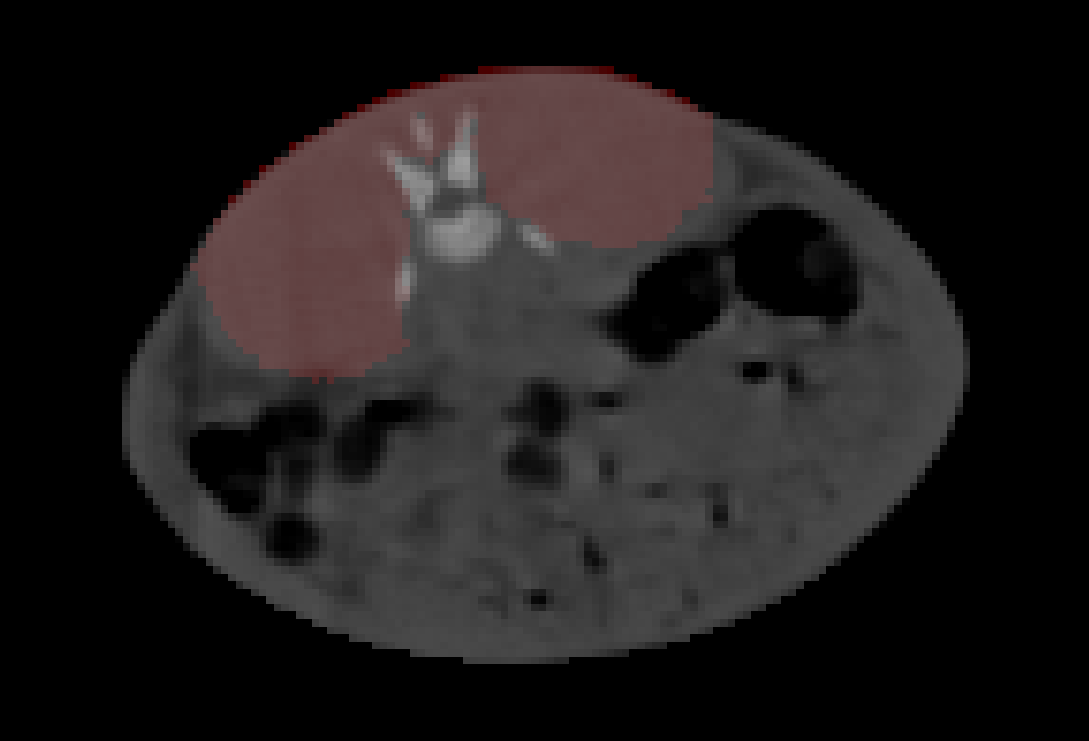}
        \caption{Cross-sectional slice at red region}
        \label{unseen-mld-masked}
    \end{subfigure}
    \begin{subfigure}[t]{0.48\linewidth}
        \centering
        \includegraphics[width=\linewidth]{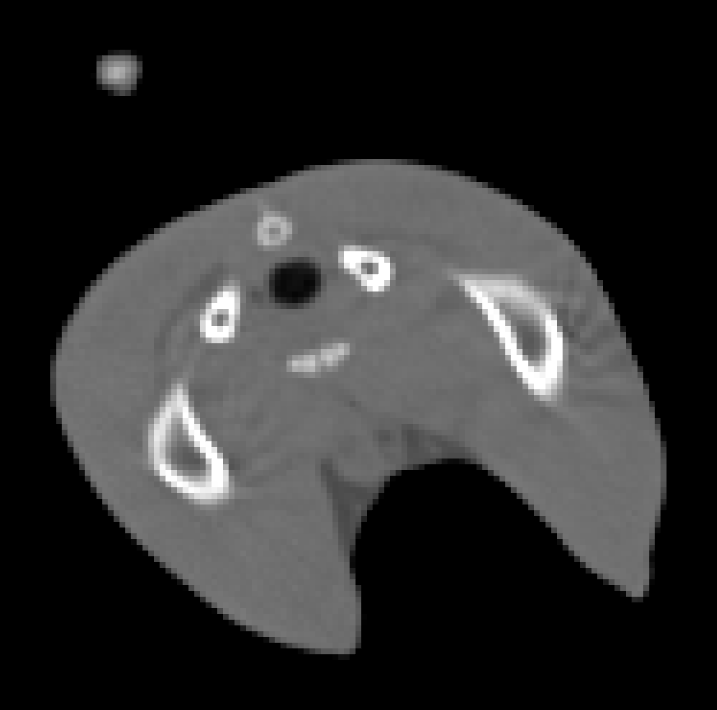}
        \caption{Cross-sectional slice at yellow region without segmentation}
        \label{unseen-hinds}
    \end{subfigure}
    \begin{subfigure}[t]{0.48\linewidth}
        \centering
        \includegraphics[width=\linewidth]{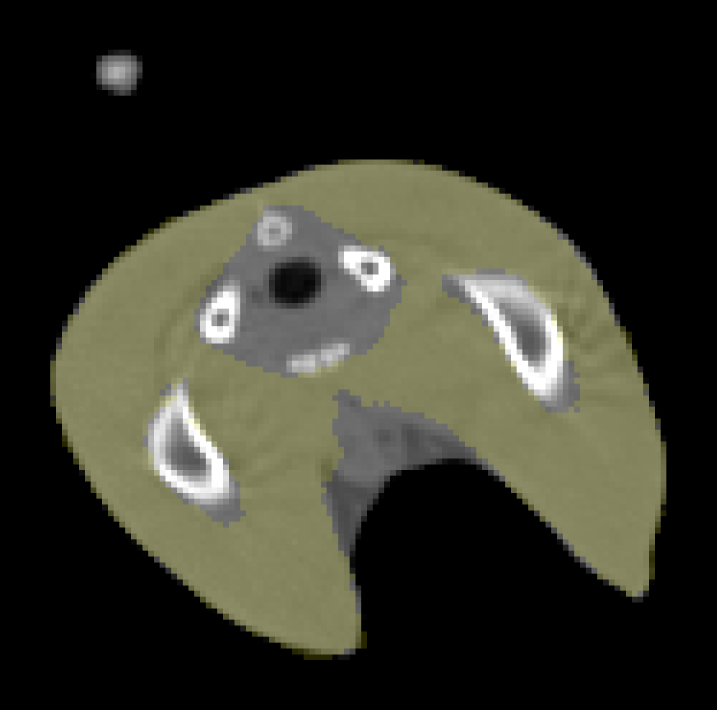}
        \caption{Cross-sectional slice at yellow region}
        \label{unseen-hinds-masked}
    \end{subfigure}
        
    \caption{Unsseen rabbits with cross-sectional slices}
    \label{fig3}
\end{figure}

For a qualitative illustration of segmentation by registration, we used the manual annotations in Figure \ref{fig1} as an atlas (A), and randomly selected the CT image of a rabbit with no manual annotations (B). Cross-sectional images in the MLD and the hind regions of rabbit B (red) are overlaid with those of atlas-rabbit A (blue) in Figure \ref{fig2} prior to and subsequent to the elastic registration. Naturally, the animals are differing in lengths and postures, thus, the initial alignment is based on their centers of mass, that is, even though the regions are roughly matching, the anatomical structures might be different at the cross-sections, which can be observed in the bone structures in Figure \ref{hinds-before-reg}. However, elastic registration transformed the body of the atlas rabbit A precisely to the body of rabbit B. In the hind region (figures \ref{hinds-before-reg}, \ref{hinds-regged}), the matching of the bone structures is spectacular. In the MLD region (figures \ref{mld-before-reg}, \ref{mld-regged}), the bone structures and the main organs and tissues are matching precisely, the only mismatch one can observe is in the visceral region, but this region is out of scope for most of the studies. The transformation field transforming the body of the atlas rabbit A to that of rabbit B can be applied to the manual segmentation mask A to fit it to the body of animal B. The result of this transformation is illustrated in Figure \ref{fig3}. As one can observe, in both cases the transformed segmentation masks fit the contours precisely and also maintain the geometrical constraints which are anchored to the body of the atlas rabbit. 
The reason why elastic registration works so accurately is the extremely strong structural information of 3D volumetric images.

In the segmentation step of the proposed method, each image in the dissected set $D$ is segmented by each atlas image from the set $M$, leading to an ensemble of segmentations for each image of $D$ with ground truth weights from dissection. 

\subsection{Feature extraction}
\label{features}

The main difference of the proposed methodology compared to human multi-atlas segmentation techniques is that our goal is not one single high quality segmentation, but the estimation of weights. Unlike in multi-atlas based segmentation techniques applied in human medicine, we do not unify the individual segmentations into one accurate one, rather, we extract statistical features to describe each segmented region and use them to build a regression model for the estimation of weights.

Since the presence of spatial patterns within tissues is limited at the resolution of the CT instruments, simple statistical descriptors of the radiodensity distributions of the segmented regions are extracted: for each segmentation in the ensemble, the total number of voxels, the total HU density, mean HU, the standard deviation, skewness, kurtosis, and the histogram of HU values are extracted, leading to a feature vector of dimensionality $(6 + H)\cdot M$, where $H$ represents the number of bins and $M$ stands for the number of atlas models used. Additionally, we extract the same features from the mean segmentation mask.
Consequently, in this step, one arrives to a regression problem of $D$ pieces of $(6 + H)\cdot (M + 1)$ dimensional feature vectors with corresponding weights measured in dissection studies.

\subsection{Model and feature selection}
\label{modelselection}

By model selection we refer to the process of finding the best regressor for a particular regression problem.
Although radiodensity is not the same as mass density and various biases might occur due to the geometry of the subject and the parameters of the acquisition, one can still expect that linear regression techniques perform well as the weights should still be approximately linear functions of the counts in the radiodensity histograms, with standard deviation, skewness and kurtosis providing possibilities for non-linear corrections. However, the relatively large number of features compared to the relatively small number of samples and the strong correlation of histogram bin counts (multicollinearity) risks overfitting, therefore, regularization is needed. 

Although some variants of linear regression incorporate regularization (like $L_1$ regularization in lasso regression \cite{bishop} and $L_2$ regularization in ridge regression \cite{bishop}), or implicitly convert the high dimensional feature vectors into a lower dimensional space (as the commonly used partial least squares (PLS) technique \cite{Font-i-Furnols2013}), we propose a more general approach: reducing the number of features by feature subset selection as part of the model selection process. As feature selection is a combinatorial optimization problem, there are no exact algorithms to solve it, however, reasonable suboptimal solutions can be found by stochastic optimization techniques, like simulated annealing \cite{sa}. Ideally, the parameters of a regressor and the ideal subset of features to operate on should be found jointly. The pseudocode of the proposed joint feature and model selection algorithm is shared in Algorithm \ref{sa-alg}: simluated annealing modifies the regressor parameters and the subset of features randomly, computes the $r^2$ scores by cross-validation and with an increasing probability accepts only those changes which improve the $r^2$ scores, thus, approaches the optimum, but still enables the discovery of the parameter space. The proposed joint feature and parameter selection technique is applicable with any regressor, in Section \ref{rabbit} we compare the linear, PLS, lasso, ridge and k-Nearest Neighbors (kNN) techniques. As a result of the joint feature and parameter selection, the ideal subset of features and regressor parameters are determined.

\begin{algorithm}
\begin{enumerate}[itemsep=0pt]
\item Set the temperature of the system to $T=1$, determine the maximum number of iterations $N$, the minimum temperature to reach $T_0$ and the annealing factor $\alpha= T_0^{\frac{1}{N}}$.
\item Select an initial subset of features and parameters for the regressor randomly. 
\item Evaluate the current configuration by computing the $E \leftarrow r^2$ score by repeated k-fold cross validation.
\item Repeat $N$ times:
\begin{enumerate}[itemsep=0pt]
\item Randomly modifiy the subset of features and the parameters of the regressor.
\item Evaluate the modified configuration by computing the $r^2$ score: $E' \leftarrow r^2$
\item With probability $exp\left(-\dfrac{E' - E}{T}\right)$ accept the new configuration and update $E \leftarrow E'$
\item Decrease the temperature of the system: $T \leftarrow \alpha T$.
\end{enumerate}
\end{enumerate}
\caption{The simulated annealing based joint feature and parameter selection}
\label{sa-alg}
\end{algorithm}

\section{Applications and results}
\label{rabbit}

In this section, we examine the suitability of the proposed approach for weight estimation in rabbit breeding selection programs (MLD region and hind fillet), and also evaluate the benefits of using multiple atlases and the importance of feature selection.

\subsection{Animals}
All animals were handled according to the principles stated in the EC Directive 86/609/EEC regarding the protection of animals used for experimental and other scientific purposes (\citep{2010/63/EU}). The study was performed on 170 Pannon White rabbits at 10 weeks old originated by the test unit of the Hungarian University of Agricultural and Life Sciences, Kaposvár Campus (former name Kaposv\'ar University) for test slaughtering and 5 rabbits for creating  multi-atlas annotations were obtained from former CT examinations. All of the animals had the same genotype and age. The test slaughtering, tissue collection and weight measurement were performed according to the recommendation of the World Rabbit Science Association (\citep{BlascoOuhayoun1996}).

\subsection{Image acquisition and preprocessing}
The image acquisition was carried out at 2018 January using a Siemens Somatom Sensation Cardiac 16 MDCT scanner as described in (\citep{Matics2020}). The CT examinations of the rabbits were performed covering the whole rabbits from head to toe using the following parameters: tube voltage 140 kV, X-ray radiation dose 90 mAs, spiral data collection mode with pitch 1, field of view 500 mm according to ISO 9001:2015 quality management system and ISO 14001:2015 environmental management system. During the CT scanning procedures 3 rabbits were immobilized in a special plastic container with belts in prone position, anaesthetics were not applied(\citep{Matics2020}). Overlapping axial scans were reconstructed from the raw data covering the whole rabbits using the Siemens Syngo CT 2007S programme with convolution kernel B65s and mediastinum window. The resolution of the images was 0.977 $\times$ 0.977 $\times$ 2 mm. The images were archived in DICOM (Digital Imaging and COmmunications in Medicine) format and then each of the series was converted to MINC metafiles (\citep{MINC}). This file type allows to be stored an acquisition process in a single file. The images were preprocessed by means of the OpenIP software package (\citep{Kovacs2010}). The individuals were separated and the plastic container holders were segmented out from the images using an automated pipeline (\citep{Kovacs2013}). For the following operations the individual images were converted to NIFTI (Neuroimaging Informatics Technology Initiative) files (\citep{NIFTI}) and the coordinate systems were set with the same starting point located at the centers of mass.
    The above described scanning and preprocessing protocol were performed on all of the examined animals.

\subsection{Experimental settings}

In both weight estimation problems, we used 5 manually segmented images, and 170 images with ground truth weights measured through dissection. 

The elastic registration by \verb|elastix| was applied with default parameters for non-rigid registration: multi-resolution fitting using b-spline transformation grid, mutual information (MI) as the similarity measure and adaptive stochastic gradient descent (SGD) to optimize the transformation field in a maximum of 200 iterations (for further details on parameters see the documentation of the elastix \cite{elastix} package). Most of the remaining parameters control various sampling rates, which can be reduced to speed up the registration. With the default settings, the registration of one atlas takes approximately 1 minute on an average desktop computer which we found acceptable and did not adjust the sampling rate parameters. Randomly checking 10\% of the fitted masks we did not find failures. A spectacular indicator of the robustness of elastic registration is that there was no need for parameter tuning, also implying that slight changes of the parameters should not change the results remarkably. 

The histograms extracted as parts of the feature vectors had the bin width of 10 HU in the range [0,200] commonly accepted as the HU range of the muscle tissue \citep{Scholz2015}. The number of bins (H=20) is aligned with the recommendation of the Sturges-rule \citep{sturges}: $H= \lceil \log_2 N\rceil + 1 = 18$ for the average size of the MLD region ($N_{MLD}$=100 000 voxels) and $H=19$ for the average size of the hind fillet ($N_{hinds}$= 200 000 voxels). Since the number of bins was not tuned, slight adjustments should not change the results remarkably.

Generally, the problem is linear: given the histogram counts $H(R)$ of a particular radiodensity $R$, the corresponding mass can be approximated with an unknown scaling factor $S(R)$ as $S(R)\cdot H(R)$. The approximate linearity of the weight estimation problem is also confirmed by the successful applications of the linear technique called partial least squares (PLS) regression \citep{Font-i-Furnols2013}. Additionally to various regularized linear techniques (lasso \citep{bishop}, ridge \citep{bishop}, PLS), we also evaluate kNN regression:, histograms (or subsets of them) being similar (close neighbors) might imply similar weights making kNN a reasonable regressor. Due to the essentially linear nature of the problem, the relatively small number of samples (170) and in accordance with the practice in the literature (using linear regression techniques dominantly), we do not consider advanced non-linear regressors (like random forests \citep{bishop}) suitable in these problems. The model selection was implemented in terms of the regressors available in the scikit-learn \cite{scikit-learn} Python package.


The parameters of the 5 reasonable regressors we optimized in the proposed joint feature subset and parameter selection technique (Algorithm \ref{sa-alg}) are summarized as follows.
\begin{itemize}[itemsep=0pt]
    \item linear regression \cite{bishop}: the only parameter being varied is whether the intersection (additional constant) is fitted;
    \item partial least squares (PLS): varying the number of components in the range $[2, 3, \dots, \sqrt{F}]$, $F$ denoting the number of features in the actual feature set;
    \item lasso regression \cite{bishop}: the $L_1$ regularized linear regression, varying the regularization factor in the range $[0.01, 10]$;
    \item ridge regression \cite{bishop}: the $L_2$ regularized linear regression, varying the regularization factor in the range $[0.01, 10]$;
    \item k nearest enighbors regression \cite{bishop} (kNN): varying the number of neighbors (1-13), the weighting being uniform or distance weighted, and the $L_p$ distance with $p\in[1, 6]$.
\end{itemize}

The joint feature and parameter selection algorithm \ref{sa-alg} was executed with a maximum of $N=8000$ iterations and $T_0=10^{-7}$ which we found reasonable for the range of the $r^2$ scores being optimized. All $r^2$ scores reported and used in the model selection and reported in the rest of the paper are based on 20 times repeated 5-fold cross-validation. For comparability, the random seeds of the cross-validation were fixed, leading to the same folds in each evaluation. With these settings, and the early stopping rule of finishing the optimization when no improvement was achieved in the last 500 iterations, the model selection takes approximately 25 minutes for all regressors together on an average desktop computer. The threshold for early stopping (500 iterations) was set according to the total number of features (156), with the interpretation that for early stopping the involvement of each feature was considered approximately 3 times in average (in various configurations) with no improvement. The early stopping rule was triggered in all test cases (usually earlier than 3000 iterations), which means that convergence is achieved at the temperature of about $T=0.002$. Since this temperature is still comparable to the differences in the $r^2$ scores one can expect from important features, the triggering of the early stopping rule was not caused by the overly decreased temperature, but the lack of improvements, thus, real convergence was achieved, which validates the numerical parameters used in the feature selection process. Nevertheless, slight changes of the parameters should not change the results remarkably.

In order to validate the concept of using features from multiple atlases, we have carried out the model selection for the features coming from the individual atlases. Furthermore, we evaluated the use of one single segmentation derived from the multiple ones by averaging (following the approach used in human medicine where the multi-atlas based segmentations are unified to one single segmentation). In order to validate the importance of feature selection, the parameters of the regressors were optimized for all features (without feature selection), as well.

\subsection{Results for the MLD weights}

\begin{table*}
\caption{The $r^2$ scores of the 20 times repeated 5-fold cross-validation on the MLD dataset, with and without feature selection. Highest values are indicated by boldface typesetting.}
\label{mld-res}
\centering
\begin{small}
\begin{tabular}{llllll}
\toprule
{} & \multicolumn{5}{l}{no feature selection} \\ \hline
{} &            linear &              PLS &            lasso &            ridge &              kNN \\
type        &                   &                  &                  &                  &                  \\
\midrule
atlas 1     &      {\bf 0.7425} $\pm$ 0.1383 &   0.7559 $\pm$ 0.143 &  {\bf 0.7655} $\pm$ 0.1325 &  {\bf 0.7591} $\pm$ 0.1359 &  {\bf 0.5881} $\pm$ 0.1351 \\
atlas 2     &      0.7304 $\pm$ 0.1617 &  0.7532 $\pm$ 0.1504 &  0.7553 $\pm$ 0.1434 &  0.7547 $\pm$ 0.1427 &    0.5419 $\pm$ 0.14 \\
atlas 3     &       0.714 $\pm$ 0.1691 &  0.7299 $\pm$ 0.1591 &    0.731 $\pm$ 0.163 &  0.7326 $\pm$ 0.1577 &  0.4965 $\pm$ 0.1244 \\
atlas 4     &      0.7008 $\pm$ 0.1564 &  0.7312 $\pm$ 0.1563 &   0.738 $\pm$ 0.1495 &  0.7292 $\pm$ 0.1525 &  0.5153 $\pm$ 0.1365 \\
atlas 5     &      0.6815 $\pm$ 0.1944 &    0.7334 $\pm$ 0.18 &  0.7373 $\pm$ 0.1642 &  0.7246 $\pm$ 0.1765 &  0.4942 $\pm$ 0.1727 \\
mean        &      0.7394 $\pm$ 0.1564 &  {\bf 0.7582} $\pm$ 0.1527 &  0.7635 $\pm$ 0.1443 &   0.758 $\pm$ 0.1485 &  0.5683 $\pm$ 0.1174 \\
multi-atlas &     -6.0739 $\pm$ 8.5356 &  0.7547 $\pm$ 0.1553 &  0.7635 $\pm$ 0.1299 &  0.7079 $\pm$ 0.1683 &  0.5698 $\pm$ 0.1208 \\
\toprule
{} & \multicolumn{5}{l}{feature selection} \\ \hline
{} &               linear &              PLS &            lasso &            ridge &              kNN \\
type        &                      &                  &                  &                  &                  \\
\midrule
atlas 1     &   0.7786 $\pm$ 0.1265 &  0.7775 $\pm$ 0.1296 &   0.7785 $\pm$ 0.126 &   0.7786 $\pm$ 0.126 &   0.733 $\pm$ 0.1178 \\
atlas 2     &    0.7728 $\pm$ 0.136 &   0.7728 $\pm$ 0.136 &  0.7706 $\pm$ 0.1358 &  0.7716 $\pm$ 0.1375 &   0.708 $\pm$ 0.1467 \\
atlas 3     &   0.7529 $\pm$ 0.1452 &  0.7454 $\pm$ 0.1541 &  0.7528 $\pm$ 0.1451 &  0.7541 $\pm$ 0.1443 &  0.6974 $\pm$ 0.1279 \\
atlas 4     &   0.7459 $\pm$ 0.1505 &  0.7456 $\pm$ 0.1459 &  0.7445 $\pm$ 0.1466 &  0.7469 $\pm$ 0.1437 &  0.6553 $\pm$ 0.1352 \\
atlas 5     &   0.7455 $\pm$ 0.1655 &   0.7577 $\pm$ 0.162 &  0.7477 $\pm$ 0.1621 &  0.7448 $\pm$ 0.1652 &  0.6937 $\pm$ 0.1314 \\
mean        &   0.7797 $\pm$ 0.1373 &  0.7788 $\pm$ 0.1431 &  0.7795 $\pm$ 0.1364 &  0.7798 $\pm$ 0.1368 &  0.7316 $\pm$ 0.1394 \\
multi-atlas &    {\bf 0.825} $\pm$ 0.1011 &   {\bf 0.808} $\pm$ 0.1127 &   {\bf 0.8273} $\pm$ 0.105 &  {\bf 0.8194} $\pm$ 0.1108 &  {\bf 0.7413} $\pm$ 0.1014 \\
\bottomrule
\end{tabular}

\end{small}

\end{table*}

\begin{figure}
\begin{center}
\includegraphics[width=0.4\textwidth]{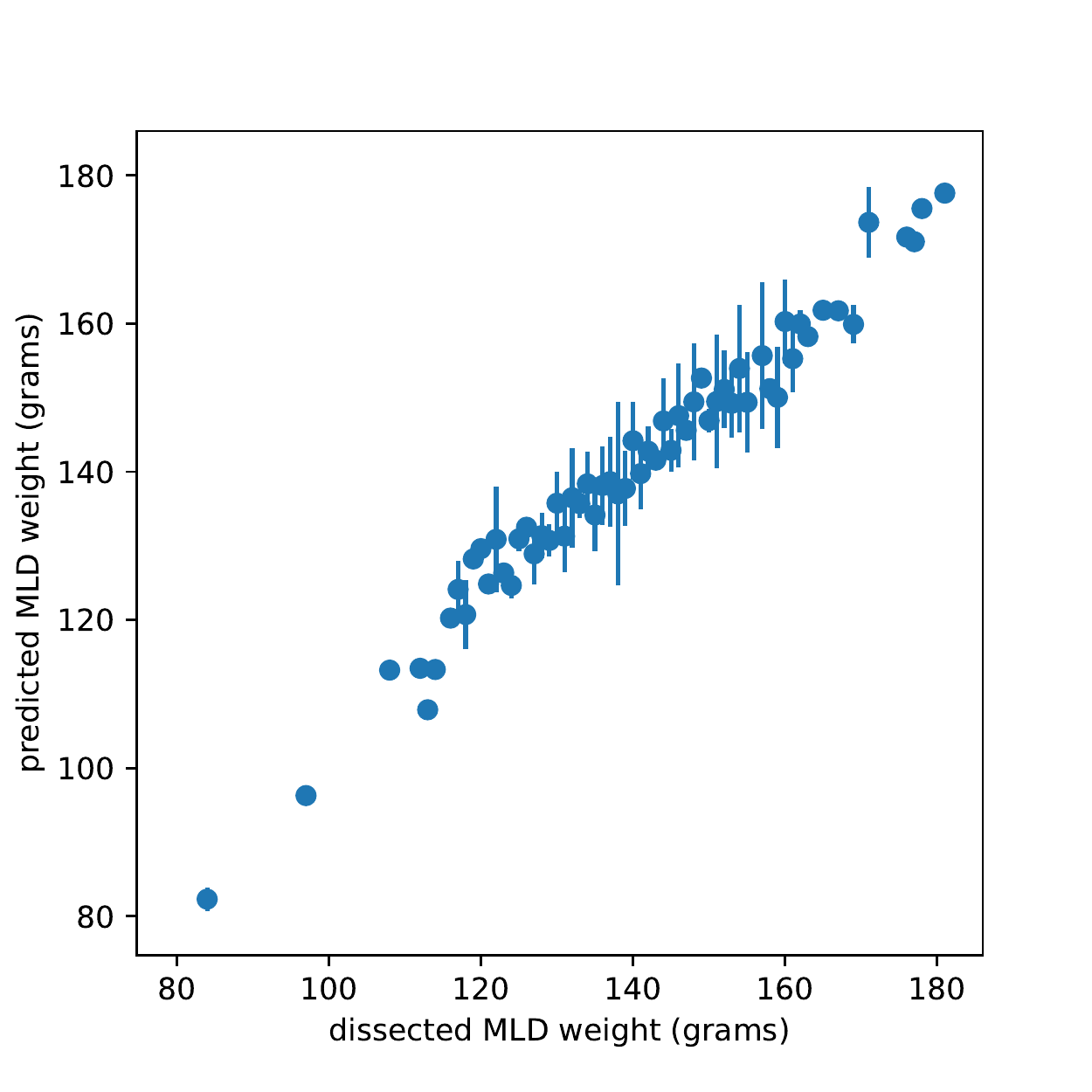}
\end{center}
\caption{Means and standard deviations of MLD predictions over the 20 times repeated 5-fold cross-validation using feature selection with multi-atlas features and lasso regression.}
\label{mld-plot}
\end{figure}

The results of the weight estimation of meat cuts in the MLD regions are summarized in Table \ref{mld-res}, with and without feature subset selection. For comparability across various problems and following the practice in the literature, we report $r^2$ scores, with the remark that $r^2$ scores are monotonic functions of the mean squared errors (mean squared residuals), thus, higher $r^2$ scores imply lower mean squared residuals. As one can observe, with all regressors, the highest scores are achieved when the multi-atlas approach is used with feature selection (lasso regression delivering the highest score overall). Furthermore, with feature selection and multi-atlas features the standard deviations of the $r^2$ scores are also remarkably lower than those of others. The relatively low scores of kNN regression compared to linear ones provide justification for the previous claims on the suitability of linear approaches for weight estimation. 

For completeness, we mention that the average MLD weight is 143 grams, and the best performing lasso technique with multi-atlas features and feature selection reached the root mean squared error of 6.5 grams. The means and standard deviations of these predictions over all cross-validation folds are plotted in Figure \ref{mld-plot}. 

Due to the multicollinearity of features (many of them are replacable to a certain extent) and the stochastic nature of the optimization, the analysis of selected feature subsets is ambiguous. However, checking the feature subsets of the best performing linear techniques, some general characteristics can be identified: the subsets are dominated by histogram features coming from all 5 atlases; features corresponding to the HU range [0-120] are rarely used; and some non-linear features (skewness or kurtosis) are always parts of the subsets.

We have applied the Wilcoxon signed-rank test (making no assumptions on the distributions) to compare the squared residuals of the multi-atlas scores to the others (atlases 1-5 and mean mask) for each regressor. All p-values are less than $0.001$, indicating that the use of features originating from multiple atlases brings a statistically significant improvement (at the usual levels of statistical significance) compared to the use of individual atlases or a unified atlas (mean mask).

Comparing the scores achieved by joint feature subset and parameter selection to the case when only parameter selection was optimized, one can observe, that feature selection has a positive effect on the results in all cases. Again, Wilcoxon signed-rank test was applied to the squared residuals of each case with and without feature selection to test statistical significance. The highest p-value of $0.0001$ suggests that feature subset selection is beneficial even in the case when only one atlas is used. One spectacular outlier is the $r^2=-6.0739$ score of linear regression with multi-atlas features, without feature selection: $r^2$ scores smaller than 0 can occur if the predictor has less predictive power than the mean of the target variable - this can happen with unregularized linear techniques due to overfitting the model to multicollinear features \citep{bishop}. The improved results of linear regression with feature selection demonstrate that feature selection acts like an additional regularization of linear techniques. Interestingly, without feature selection, in multiple cases the highest scores were achieved by feature sets from single atlases. This result confirms the importance of feature subset selection: due to the small number of samples and high number of features, even the regularized linear techniques (PLS, lasso, ridge) fail to generalize well when too many features are used. 

\begin{table*}
\caption{The $r^2$ scores of the 20 times repeated 5-fold cross-validation on the hinds fillet dataset, with and without feature selection. Highest values are indicated by boldface typesetting.}
\label{hinds-res}

\centering
\begin{small}
\begin{tabular}{llllll}
\toprule
{} & \multicolumn{5}{l}{no feature selection} \\ \hline
{} &               linear &              PLS &            lasso &            ridge &              kNN \\
type        &                      &                  &                  &                  &                  \\
\midrule
atlas 1     &       {\bf 0.8647} $\pm$ 0.089 &  {\bf 0.8732} $\pm$ 0.0843 &  {\bf 0.8759} $\pm$ 0.0815 &   {\bf 0.875} $\pm$ 0.0805 &  0.6405 $\pm$ 0.1374 \\
atlas 2     &      0.8489 $\pm$ 0.0953 &  0.8683 $\pm$ 0.0873 &  0.8704 $\pm$ 0.0832 &  0.8653 $\pm$ 0.0864 &  0.6779 $\pm$ 0.1404 \\
atlas 3     &      0.8524 $\pm$ 0.1004 &   0.868 $\pm$ 0.0867 &  0.8717 $\pm$ 0.0829 &  0.8655 $\pm$ 0.0885 &  0.6092 $\pm$ 0.1159 \\
atlas 4     &      0.8545 $\pm$ 0.0924 &  0.8712 $\pm$ 0.0891 &  0.8729 $\pm$ 0.0839 &   0.8688 $\pm$ 0.084 &  0.6572 $\pm$ 0.1332 \\
atlas 5     &       0.848 $\pm$ 0.1038 &  0.8661 $\pm$ 0.0887 &  0.8719 $\pm$ 0.0822 &  0.8623 $\pm$ 0.0898 &  {\bf 0.7095} $\pm$ 0.1277 \\
mean        &      0.8521 $\pm$ 0.1035 &  0.8712 $\pm$ 0.0892 &  0.8745 $\pm$ 0.0828 &  0.8701 $\pm$ 0.0881 &  0.6354 $\pm$ 0.1348 \\
multi-atlas &     -1.7193 $\pm$ 3.7919 &  0.8693 $\pm$ 0.0897 &  0.8625 $\pm$ 0.0845 &  0.8403 $\pm$ 0.1009 &  0.6775 $\pm$ 0.1389 \\

\bottomrule
{} & \multicolumn{5}{l}{feature selection} \\ \hline
{} &            linear &              PLS &            lasso &            ridge &              kNN \\
type        &                   &                  &                  &                  &                  \\
\midrule
atlas 1     &   0.8807 $\pm$ 0.0783 &  0.8808 $\pm$ 0.0809 &  0.8857 $\pm$ 0.0743 &   0.8845 $\pm$ 0.076 &  {\bf 0.8361} $\pm$ 0.1108 \\
atlas 2     &   0.8755 $\pm$ 0.0815 &  0.8783 $\pm$ 0.0772 &   0.879 $\pm$ 0.0788 &  0.8792 $\pm$ 0.0779 &   0.827 $\pm$ 0.1005 \\
atlas 3     &      0.87 $\pm$ 0.081 &  0.8812 $\pm$ 0.0778 &    0.8785 $\pm$ 0.08 &  0.8808 $\pm$ 0.0769 &  0.8197 $\pm$ 0.0999 \\
atlas 4     &   0.8721 $\pm$ 0.0853 &   0.8815 $\pm$ 0.078 &  0.8786 $\pm$ 0.0823 &  0.8719 $\pm$ 0.0858 &  0.8136 $\pm$ 0.0976 \\
atlas 5     &     0.8773 $\pm$ 0.08 &  0.8764 $\pm$ 0.0868 &    0.88 $\pm$ 0.0769 &   0.878 $\pm$ 0.0777 &   0.8274 $\pm$ 0.116 \\
mean        &   0.8789 $\pm$ 0.0818 &   0.883 $\pm$ 0.0788 &  0.8809 $\pm$ 0.0792 &  0.8816 $\pm$ 0.0807 &  0.8246 $\pm$ 0.0887 \\
multi-atlas &    {\bf 0.897} $\pm$ 0.0769 &   {\bf 0.8872} $\pm$ 0.074 &  {\bf 0.8902} $\pm$ 0.0752 &  {\bf 0.8988} $\pm$ 0.0658 &  0.8275 $\pm$ 0.0949 \\
\bottomrule
\end{tabular}
\end{small}
\end{table*}

The results confirm that the use of multi-atlas features with feature subset selection and linear techniques are a suitable choice for weight estimation. In order to see if there was a significant difference between linear techniques, again, we exploited Wilcoxon singed-rank tests on the squared residuals, and found no statistically significant difference between the linear, lasso and ridge techniques.

\subsection{Results for the hinds fillet weights}

The analysis we carried out to estimate the weights of the hinds fillet region is structured similarly to that of the MLD region, with the $r^2$ scores summarized in Table \ref{hinds-res}. Again, the linear techniques with multi-atlas features and feature selection outperform single atlas feature sets with statistical significance in all cases (with a maximum Wilcoxon p-value of $0.001$ regarding the squared residuals). Similarly to the case of MLD, one can observe a decrease in the standard deviations using multi-atlas features with feature selection. However, in this case ridge regression provided the highest scores.

\begin{figure}
\begin{center}
\includegraphics[width=0.4\textwidth]{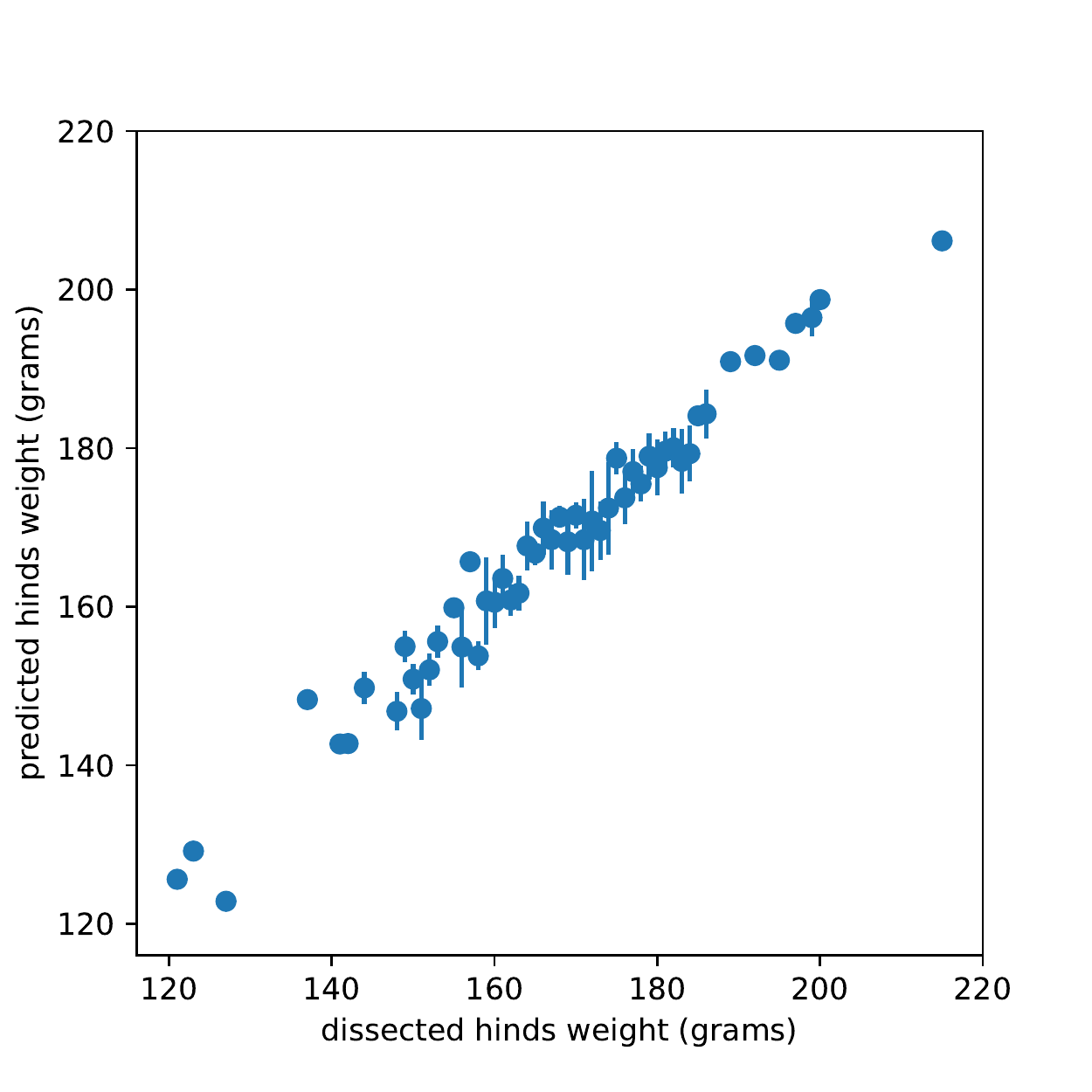}
\end{center}
\caption{Means and standard deviations of hind fillet predictions over the 20 times repeated 5-fold cross-validation using feature selection with multi-atlas features and ridge regression.}
\label{hinds-plot}
\end{figure}

The average weight of hind fillets is 334 grams, and the best performing ridge technique achieved the root mean squared error of 8.8 grams in 20 times repeated 5-fold cross validation. The means and standard deviations of these predictions over all cross-validation folds are plotted in Figure \ref{hinds-plot}. Regarding the selected feature subsets, we found that features from all atlases are used, and no HU range is excluded so remarkably as in the case of the MLD. Interestingly, there is an absense of non-linear descriptors like skewness and kurtoses. This can be explained by the following reasons. On the one hand, the hinds fillet is larger than the MLD, leading to more reliable histograms. On the other hand, the absence of global non-linear descriptors suggests that the bias of the dissection (affecting the distributions globally) is less dominant than in the MLD case. This can also be confirmed qualitatively by observing the complex structure of the backbone with spikes in the MLD region, which can be accountable for the decreased precision of dissection.

The comparison of scores with and without feature selection shows similar patterns as with the MLD weight estimation: in all cases, the scores with feature selection are higher than without feature selection. The highest Wilcoxon p-value of $0.00001$ regarding the squared residuals indicates statistical significance in all cases except one outlier: ridge regression with atlas 4 has the p-value of $0.2$, in this case the improvements by feature selection are not significant.

Finally, comparing the results of the various linear techniques when multi-atlas features are used with feature selection, the differences observable in Table \ref{hinds-res} are statistically significant with the highest Wilcoxon p-value of $10^{-7}$ regarding squared residuals, except linear and ridge regression, the difference between their $r^2$ scores $0.897$ and $0.8988$ is not significant. The results of the two experiments suggest that linear and ridge regression with feature selection are the most suitable techniques in this class of problems.

\subsection{Comparison to dedicated segmentation solutions on rabbits}

In a previous paper of ours \citep{Matics2020}, a dedicated segmentation method was developed for the MLD region and evaluated on the same database, resulting $r^2=0.74$. The proposed technique reached $r^2=0.83$ in the MLD region, an improvement of 0.08 on the absolute scale, and 12\% relative to the base score of 0.74. Another former study reports the correlation coefficienct r=0.89 ($r^2=0.79$) between the CT volume of the thigh muscles and its weight \citep{Gyovai2013}). On the hinds fillet, the proposed method reaches the $r^2=0.89$, which is an improvement of 0.1 on the absolute scale, and 12\% relative to the base score of 0.79. We highlight that in the latter case the scores are computed on different images, however, both studies serve the same purpose, so the comparison is meaningful. 

Finally, we mention that the selection procedure of Pannon White rabbits was based on the method \citep{Nagy2006} from 2004 until now. Due to the improvements in prediction accuracy, the developed methods provide the basis for the ongoing breeding selection program of Pannon White rabbits at the Hungarian University of Agricultural and Life Sciences, Kaposvár Campus.

\section{Summary}
\label{conclusions}
 
In this paper, we proposed a general approach for problems in farm animal imaging where the estimation of weights of body parts of (possibly) living animals is needed (Section \ref{prob}). The proposed method adapts the multi-atlas based segmentation developed in human medicine, driven by elastic registration to handle the high variability of the subjects (Section \ref{segreg}). Compared to multi-atlas techniques in medicine, the main difference is that we do not unify the individual segmentations into one high quality segmentation, but feed the statistical descriptors of the ensemble of segmentations into the subsequent regression step (Section \ref{features}). In order to cape with the relatively large number of features and relatively low number of samples, the regression is optimized in a joint feature and parameter selection procedure (Section \ref{modelselection}). The proposed solution fulfills the requirements phrased in Section \ref{prob}: it is fully automated, enabling the processing of CT scans at scale; able to handle the high variability of subjects due to elastic registration; the use of manual annotations (atlases) enables the declaration of any geometrical and anatomical constraints; finally, the adaptation to a new problem requires only a handful of manual segmentations and a dissection study. 

The method is evaluated in two problems estimating the weights of valuable meat cuts in rabbits for breeding selection programs (Section \ref{real}). In both problems, and achieved 12\% higher $r^2$ scores than previous, dedicated segmentation based solutions, without the need of any problem specific development (Section \ref{rabbit}). The improved $r^2$ scores lead to the replacement of the currently used approach \citep{Nagy2006} to the proposed weight estimation technique in the breeding selection program of Pannon White rabbits at the Hungarian University of Agricultural and Life Sciences, Kaposvár Campus.

The proposed method was also evaluated without the two important ingredients: the use of features from multiple atlases and the application of feature selection. On both datasets, for all variations of linear regression, statistical hypothesis tests show statistical significance regarding the benefits of using features from multiple atlases and carrying out feature selection. Regarding the regression techniques, no significant difference was found between linear regression and the best performing ridge regression.

Due to the generality of the method, further steps include its application to other farm animal breeding programs such as broiler chicken, turkey, goose, pig and a retrospective evaluation of former CT images. 

The multi-atlas based segmentation, feature extraction and model selection steps are released as an open source Python package (\url{http://github.com/gykovacs/maweight}) for the benefit of the community. Furthermore, for complete reproducibility, all CT images, manual segmentations and evaluation notebooks are shared in the GitHub repository \url{http://github.com/gykovacs/rabbit_ct_weights}.

\section*{Author Contributions}
Conceptualization, G.K., T.D.; formal analysis, \'A.C., G.K.; funding acquisition, T.D.; investigation, V.\'A., T.D., \'A.C., I.N. and Z.G.; methodology, G.K., T.D., Z.M., \"O.P., Z.S.; project administration, \'A.C., T.D. and Z.M.; ensuring tomographic background, I.R. and M.M.; software, \'A.C. and G.K.; writing original draft, \'A.C., G.K., T.D.; writing review and editing, G.K., \'A.C., T.D. and Z.M. All authors have read and agreed to the published version of the manuscript.

\section*{Funding}
The publication is supported by the János Bolyai Research Scholarship of the Hungarian Academy of Sciences (BO/00871/19, BO/00921/19) and EFOP-3.6.3-Vekop-16-2017-00005 project. The project is co-financed by the European Union and the European Social Fund.

\section*{Acknowledgments}
The authors would like to thank all the colleagues who contributed to the experiment, especially Roz\'alia Kasza, Istv\'an Radnai.

\section*{Conflicts of Interest}
The authors declare no conflict of interest.

\section*{References}
\bibliography{refs.bib}

\end{document}